\documentclass[10pt,twocolumn,letterpaper]{article}

\usepackage{wacv}
\usepackage{times}
\usepackage{epsfig}
\usepackage{graphicx}
\usepackage{amsmath}
\usepackage{amssymb}

\usepackage{tabularx}
\usepackage{pifont}
\usepackage{adjustbox}
\usepackage{multirow}
\usepackage[caption=false,font=footnotesize]{subfig}
\usepackage{amsmath,scalerel}
\usepackage[accsupp]{axessibility}  

\def\wacvPaperID{769} 

\wacvfinalcopy 

\ifwacvfinal
\fi

\ifwacvfinal
\usepackage[breaklinks=true,bookmarks=false]{hyperref}
\else
\usepackage[pagebackref=true,breaklinks=true,colorlinks,bookmarks=false]{hyperref}
\fi

\begin{document}

\title{FastAno: Fast Anomaly Detection\\via Spatio-temporal Patch Transformation}

\author{
	Chaewon Park \quad
	MyeongAh Cho \quad
	Minhyeok Lee \quad
	Sangyoun Lee
	\thanks{Corresponding Author. \newline \hspace*{0.16in} This work was supported by the Institute of Information \& communications Technology Planning \& Evaluation(IITP) grant funded by the Korea government(MSIT) (No. 2021-0-00172, The development of human Re-identification and masked face recognition based on CCTV camera)}
	\vspace{0.01cm}\\
	Yonsei University, Seoul, Republic of Korea \\
	{\tt\small \{chaewon28, maycho0305, hydragon516,syleee\}@yonsei.ac.kr}
}
\setlength{\textfloatsep}{7pt}
\maketitle

\ifwacvfinal
\thispagestyle{empty}
\fi

\begin{abstract}
	Video anomaly detection has gained significant attention due to the increasing requirements of automatic monitoring for surveillance videos. Especially, the prediction based approach is one of the most studied methods to detect anomalies by predicting frames that include abnormal events in the test set after learning with the normal frames of the training set. 
	However, a lot of prediction networks are computationally expensive owing to the use of pre-trained optical flow networks, or fail to detect abnormal situations because of their strong generative ability to predict even the anomalies. 
	To address these shortcomings, we propose spatial rotation transformation (SRT) and temporal mixing transformation (TMT) to generate irregular patch cuboids within normal frame cuboids in order to enhance the learning of normal features.  Additionally, the proposed patch transformation is used only during the training phase, allowing our model to detect abnormal frames at fast speed during inference.   
	Our model is evaluated on three anomaly detection benchmarks, achieving competitive accuracy and surpassing all the previous works in terms of speed.

\end{abstract}

\thispagestyle{empty}


\begin{figure}[!t]
	\begin{center}
		\includegraphics[width=1.0\columnwidth]{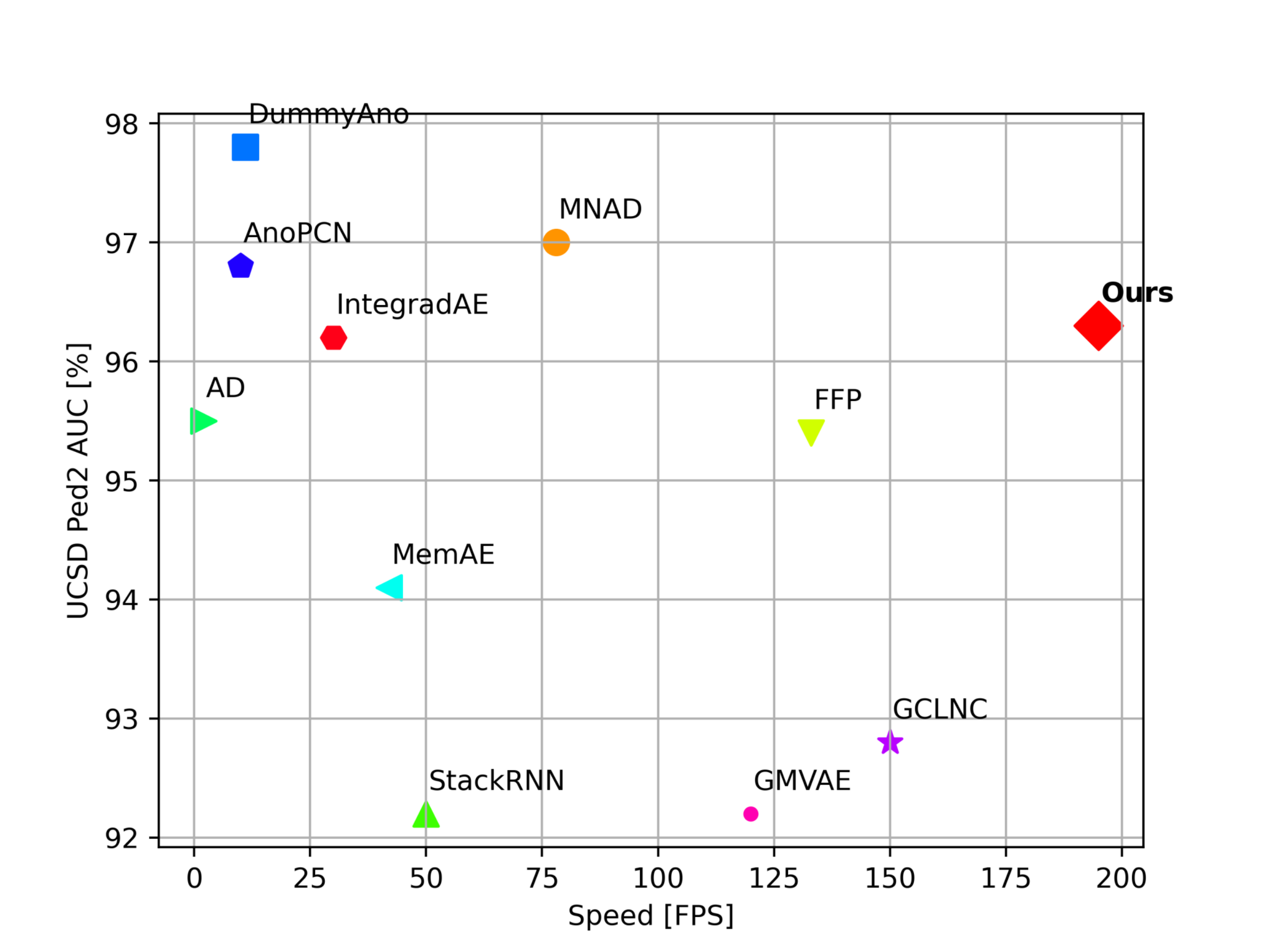}
	\end{center}
	\caption{Comparison of evaluation speed (FPS) and frame-level AUC (\%) in Ped2 test set. The methods compared in this figure are listed in Table~\ref{score}. Our framework demonstrates state-of-the-art in terms of FPS and performs competitively with other methods.}
	\label{speed}
\end{figure}
\vspace{-1.0em}

\vspace{-0.2cm}

\section{Introduction}

Video anomaly detection refers to the task of recognizing unusual events in videos. It has gained attention due to the implementation of video surveillance systems. Surveillance cameras are widely used for public safety. However, the monitoring capacity is not up to the mark. Since abnormal events rarely happen in the real world compared to normal events, automatic anomaly detection systems are in high demand to reduce the monitoring burden. However, it is very challenging because obtaining the datasets is difficult owing to the imbalance of events and variable definitions of abnormal events based on the context of each video. \\
\indent One of the challenging factors of anomaly detection is the data imbalance problem, meaning that the abnormal scenes are more difficult to capture than normal scenes because of their scarcity in the real world. Therefore, datasets with an equal number of both types of scenes are hard to obtain, and consequently, only the normal videos are provided as training data~\cite{10.1145/1541880.1541882}. This is known as an unsupervised approach for anomaly detection used by most of the previous works~\cite{hasan2016learning, Nguyen_2019_ICCV,  ravanbakhsh2017abnormal, ravanbakhsh2019training}. The unsupervised network needs to learn the representative features of the unlabeled normal training set and sort the frames with outlying features to detect abnormal events. Autoencoder (AE)~\cite{hinton2006reducing}-based methods~\cite{Abati_2019_CVPR, zaheer2020old, Liu_2018_CVPR} have proven to be successful for such a task. Frame predicting AEs~\cite{Liu_2018_CVPR, tang2020integrating} and frame reconstructing AEs~\cite{nguyen2019hybrid, hasan2016learning} have been proposed assuming that anomalies that are unseen in the training phase cannot be predicted or reconstructed when the model is trained only on normal frames. However, these methods do not consider the drawback of AE—that AE may generate anomalies as clearly as normal events due to the strong generalizing capacity of convolutional neural networks (CNNs)~\cite{gong2019memorizing}. To minimize this factor, Gong~\etal~\cite{gong2019memorizing} and Park~\etal~\cite{park2020learning} proposed memory-based methods to use only the most essential features of normal frames for the generation. However, the memory-based methods are not efficient for videos with various scenes because their performance is highly dependent on the number of items. Many memory items are required to read and update patterns of various scenes, slowing down the detection.\\
\indent Another critical and challenging issue for video anomaly detection is the performing speed. The main purpose of anomaly detection is to detect abnormal events or emergencies immediately, but slow models do not meet this purpose. In the previous studies, the following factors are observed to slow down the detection speed: heavy pre-trained networks such as optical flow~\cite{Liu_2018_CVPR, ravanbakhsh2017abnormal, ravanbakhsh2019training, yu2020cloze}, object detectors~\cite{georgescu2021anomaly, ionescu2019object}, and pre-trained feature extractors~\cite{ravanbakhsh2018plug, sultani2018real}. These modules are complex and computationally expensive. 

\indent Therefore, we take the detection speed into account and employ a patch transformation method that is used only during training. We implement this approach by artificially generating abnormal patches via applying transformations to patches randomly selected from the training dataset. We adopt spatial rotation transformation~(SRT) and temporal mixing transformation~(TMT) to generate a patch anomaly at a random location within a stacked frame cuboid. Given this anomaly-included frame cuboid, our AE is trained to learn the employed transformation and predict the upcoming normal frame. The purpose of SRT is to generate an abnormal appearance and encourage the model to learn spatially invariant features of normal events. For instance, when a dataset defines walking pedestrians as normal and all others as abnormal, by giving a sequence of a rotated person ({\it e.g.,} upside-down, lying flat) and forcing the model to generate a normally standing person, the model learns normal patterns of pedestrians. TMT, which is shuffling the selected patch cube in the temporal axis to create abnormal motion, is intended to enhance learning temporally invariant features of normal events. Given a set of frames where an irregular motion takes place in a small area, the model has to learn how to rearrange the shuffled sequence in the right order to correctly predict the upcoming frame. \\
\indent To the best of our knowledge, unlike~\cite{Liu_2018_CVPR,gong2019memorizing, park2020learning, ravanbakhsh2017abnormal, ionescu2019object, ravanbakhsh2018plug}, our framework performs the fastest because there are no additional modules or pre-trained networks. Furthermore, the proposed patch transformation does not drop the speed because it is detached during detection. Likewise, we designed all components of our method considering the detection speed in an effort to make it suitable for anomaly detection in the real world.\\
\indent We summarize our contributions as follows: 

\begin{itemize}
	\item We apply a patch anomaly generation phase to the training data to enforce normal pattern learning, especially in terms of appearance and motion.
	\item The proposed patch generation approach can be implemented in conjunction with any backbone network during the training phase.
	\item  Our model performs at very high speed and at the same time achieves competitive performance on three benchmark datasets without any pre-trained modules ({\it e.g,} optical flow networks, object detectors, and pre-trained feature extractors).
\end{itemize}

\begin{figure*}[!ht]
	\begin{center}
		\includegraphics[width=0.9\linewidth]{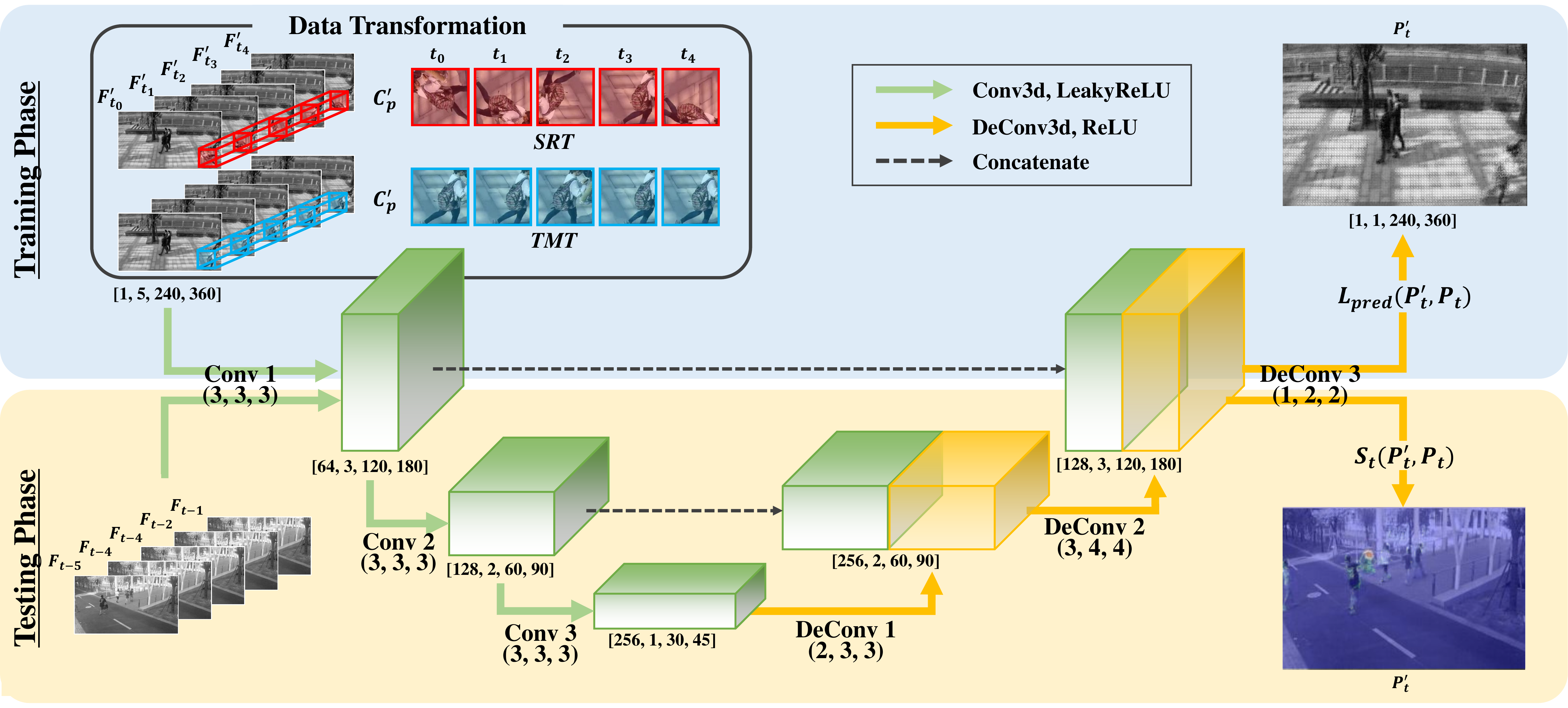}
	\end{center}
	\caption{The overview of our framework. During the training phase, SRT and TMT are employed to make our input $\mathbf{C'_f}$. The AE is trained to generate a succeeding frame that mimics the normal frame. During the testing phase, frames are fed into the AE and the corresponding output $\mathbf{P'_t}$ is generated. The normality score $S_t(P'_t, P_t)$ is used to discriminate abnormal frames. The $\mathbf{P'_t}$ in this figure is a combination of $\mathbf{P'_t}$ and a difference map for better understanding. The values in brackets indicate [channel, temporal, height, width] of feature and  (depth, height, width) of the kernel in order. }
	\label{modelarchitecture}
\end{figure*}

\vspace{-0.3em}

\vspace{-0.9em}
\section{Related work}
\subsection{AE-based approach}
Frame predicting and reconstructing AEs have been proposed under the assumption that models trained only on normal data are not capable of predicting or reconstructing abnormal frames, because these are unseen during training. Some studies~\cite{Liu_2018_CVPR, tang2020integrating, park2020learning, lu2019future} trained AEs that predict a single future frame from several successive input frames. Additionally, many effective reconstructing AEs~\cite{nguyen2019hybrid, zaheer2020old, park2020learning, cho2020unsupervised} have been proposed. Cho~\etal~\cite{cho2020unsupervised} proposed two-path AE, where two encoders were used to model appearance and motion features. Focusing on the fact that abnormal events occur in small regions, patch-based AEs~\cite{zaheer2020old, xu2017detecting, nguyen2019hybrid, fan2020video}, have been proposed. However, it has been observed that AEs tend to generalize well to generate abnormal events strongly, mainly due to the capacity of CNNs, which leads to missing out on anomalies during detection. To alleviate this drawback, Gong~\etal~\cite{gong2019memorizing} and Park~\etal~\cite{park2020learning} suggested networks that employ memory modules to read and update memory items. These methods showcased outstanding performance on several benchmarks. However, they are observed to be ineffective for large datasets due to the limitation of memory size. Furthermore, some works~\cite{Liu_2018_CVPR, ravanbakhsh2017abnormal, ravanbakhsh2019training, ravanbakhsh2018plug} have used optical flow to estimate motion features because information of temporal patterns is crucial in anomaly detection.  

\subsection{Transformation-based approach} 
Many image transformation methods, such as augmentations, have been proposed to increase recognition performance and robustness in varying environments in limited training datasets. This technique was first applied to image recognition and was later extended to video recognition. 
For the image-level modeling, Komodakis~\etal~\cite{komodakis:hal-01832768} suggested unsupervised learning for image classification by predicting the direction of the rotated input.  Krizhevsky~\etal~\cite{krizhevsky2012imagenet} used rotation, flipping, cropping, and color jittering to enhance learning spatially invariant features. Furthermore, DeVries~\etal~\cite{devries2017improved} devised CutOut, a method that deletes a box at a random location to prevent the model from focusing only on the most discriminative regions. Zhang~\etal~\cite{zhang2017mixup} proposed MixUp which blends two training data on both the images and the labels. Yun~\etal~\cite{yun2019cutmix} put forth a combination of CutOut and MixUp, called CutMix. 
For the video-level model, augmentation techniques have been extended to the temporal axis. Ji~\etal~\cite{ji2019learning} proposed a method called time warping and time masking, which randomly skips or adjusts temporal frames. \\
\indent Several studies have used the techniques mentioned above for video anomaly detection based on the assumption that applying transformations to the input forces the network to embed critical information better. Zaheer~\etal~\cite{zaheer2020old} suggested a pseudo anomaly module to create an artificial anomaly patch by blending two arbitrary patches from normal frames. They reconstructed both normal and abnormal patches and trained a discriminator to predict the source of the reconstructed output. Hasan~\etal~\cite{hasan2016learning} and Zhao~\etal~\cite{zhao2017spatio} sampled the training data by skipping a fixed number of frames in the temporal axis. Moreover, Joshi~\etal~\cite{joshi2019unsupervised} generated abnormal frames from normal frames by cropping an object detected with a semantic segmentation model and placing it in another region in the frame to generate an abnormal appearance. Wang~\etal~\cite{wang2020cluster} applied random cropping, flipping, color distortion, rotation, and grayscale to the entire frame. In contrast to these methods, our network embeds normal patterns by training from frames with anomaly-like patches. We transform the input frames along the spatial axis or temporal axis to generate abnormal frames within training datasets. Georgescu~\etal~\cite{georgescu2021anomaly} used sequence-reversed frames for a self-supervised binary classification task where the network guesses whether the given samples are regular or not. On the other hand, our method predicts the original frame from sequence transformed input and learns the normal patterns.

\vspace{-0.2cm}
\section{Proposed approach}
This section presents an explicit description of our model formation. Our model consists of two main phases: (1) the patch anomaly generation phase and (2) the prediction phase.
\vspace{-0.3em}
\subsection{Overall architecture}
Fig.~\ref{modelarchitecture} presents the overview of our framework. During the training phase, we first load $n$ adjacent frames to make a frame cuboid. After that, we apply our patch anomaly generation to the frame cuboid, which is forwarded to the AE. Our AE extracts spatial and temporal patterns of the input and generates a future frame. During inference, the patch anomaly generation is not employed. A raw frame cuboid is fed as an input to the AE. The difference between the output of the AE and the ground truth frame is used as a score to judge normality.
\vspace{-0.3em}
\subsection{Patch anomaly generation phase}
\label{section3}
Abnormal events in videos are categorized into two large branches: (1) anomalies regarding appearances ({\it e.g.,} pedestrians on a vehicle road) and (2) anomalies regarding motion ({\it e.g.,} an illegal U-turn or fighting in public). Hence, it is important to learn both the appearance and motion features of normal situations to detect anomalies in both cases.  \\
\indent The patch anomaly generation phase takes place before feeding the frames to the generator. We load $n$ successive frames $(\mathbf{F_t, F_{t+1}, F_{t+2},} \dots , \mathbf{F_{t+n-1}})$, resize each to $240 \times 360$, and concatenate them on the temporal axis to form a 4D cuboid $\mathbf{C_f} \in \mathbb{R}^{C \times n \times 240 \times 360}$, where $C$ denotes the number of channels for each frame. After that, we select a patch cuboid $\mathbf{C_p} \in \mathbb{R}^{C \times n \times 60 \times 60}$ from a random location within $\mathbf{C_f}$ to apply transformation. Since anomalies usually occur in foregrounds, we exclude a margin of 12.5 percent in length from the top and bottom of the width of $\mathbf{C_f}$ from the selection area. We heuristically find that these marginal regions are generally backgrounds. Therefore, they commonly do not contain moving objects. Thus, by limiting the range, $\mathbf{C_p}$ is more likely to capture foregrounds than backgrounds, encouraging the model to concentrate on the moving objects. Then we apply SRT or TMT to $\mathbf{C_p}$ to form a transformed patch cuboid $\mathbf{C'_p}$. Only one of the two is applied randomly for every input. \\
\indent For SRT, each patch is rotated in a random direction between 0\textdegree, 90\textdegree, 180\textdegree, and 270\textdegree, following the approach of~\cite{komodakis:hal-01832768}. By forwarding these transformed frame cuboids $\mathbf{C'_f}$ $(\mathbf{F'_t, F'_{t+1}, F'_{t+2},} ... , \mathbf{F'_{t+n-1}})$ to the frame generator, our network is encouraged to focus on the abnormal region and recognize the spatial features of the normal appearances. Suppose a network is being trained on a dataset of people walking on a road. When it is given a frame cuboid with an upside-down person created by 180\textdegree~rotation among all the other normal pedestrians and is programmed to predict a next normal scene, the network would learn the spatial features of a normal person, such as the head and the feet are generally placed at the top and bottom, respectively. Our SRT is demonstrated as follows:
\vspace{-0.2cm}
\begin{equation}
	SRT(\mathbf{F_i})=R(\mathbf{F} _ {{ \mathbf{i} }_{(x, y) \in [(x, x+W _ { p } ), (y, y+H _ { p } )]}}, \delta_i ), 
\end{equation}
\vspace{-0.5cm}

where $R$ represents the rotation function for a patch within the pixel range of $[x, x+W _ { p } ]$ in the width axis and $[y, y+H _ { p } ]$ in the height axis of input frame $F_i$. $\delta_i$ denotes the randomly set direction for the ${ i ^ { th } }$ frame, where $i$ is the index of the input frame in the range $[0, n-1]$. Furthermore, $W_{p}$ and $H_{p}$ represent the fixed width and height of the patch, respectively. The final $\mathbf{C'_f}$ is generated by concatenating the transformed $\mathbf{F'_i}$ in the temporal axis. \\
\indent TMT involves shuffling the sequence of the patch cuboid $\mathbf{C_p}$ in the temporal axis with the intention of generating abnormal movement. The network needs to detect the awkward motion and match the sequence to normal before predicting the next frame to reduce the loss and generate a frame as similar as possible to the ground truth. For example, when the patch sequence is reversed, and a backward-walking person is generated within a frame where only forward walking people are annotated as normal, the model should find the correct sequence of the abnormal person based on the learned features to predict the correct trajectory. Our TMT function is as follows:
\begin{equation}
	TMT(\mathbf{F_i})=T(\mathbf{F _ { i }} , \mathbf{F} _ {\mathbf{\xi}_{{{ i }_{(x, y) \in [(x, x+W _ { p } ), (y, y+H _ { p } )]}}}}),
\end{equation}
where $T$ denotes a function that copies a patch located in pixel range of $[x, x+W _ { p } ]$ in the width axis and $[y, y+H _ { p } ]$ in the height axis of input frame $\mathbf{F _ { \xi_{i} }}$ and pastes it to the $i ^ { th }$ frame.  $\xi$ represents the shuffled sequence of $n$ patches ({\it e.g.} sequence $(4, 1, 0, 3, 2)$ when $n$ is 5). Same as SRT, the final $C'_f$ is the stack of the transformed $F_i$.

\indent Our patch anomaly generation phase is computationally cheaper than the other methods that embed spatio-temporal feature extraction in networks, such as storing and updating memory items~\cite{gong2019memorizing, park2020learning}, and estimating optical flow with pre-trained networks~\cite{Liu_2018_CVPR, ravanbakhsh2017abnormal, yu2020cloze, cai2021appearance}. Therefore, our patch anomaly generation phase boosts feature learning at a low cost. Furthermore, this phase is not used during the inference, meaning that it does not affect the detection speed at all. Thus, our model is low in complexity and computational costs (see Section~\ref{section4}).

\begin{figure}[t]
	\centering
	\subfloat[SRT]{\includegraphics[width=1\linewidth]{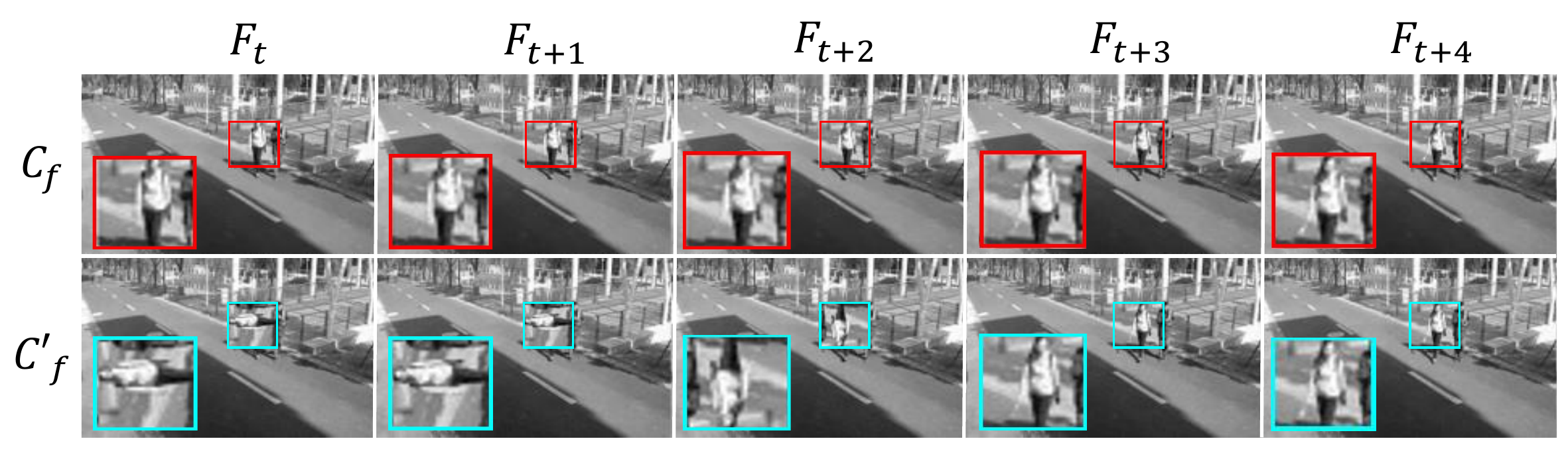}
	}
	\\
	\vspace{-0.8em}
	\subfloat[TMT]{\includegraphics[width=1\linewidth]{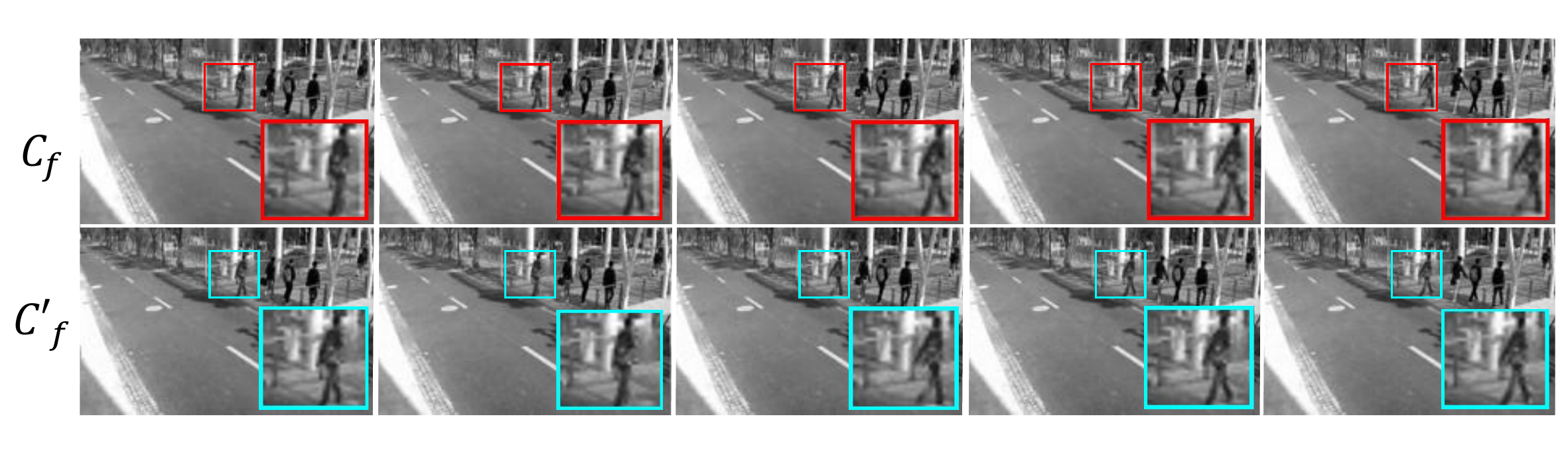}
	}\\
	\vspace{0.5em}
	\caption{Visualization of (a) SRT and (b) TMT. The frames in the upper rows are components of $C_f$. The regions marked in color are the locations of the selected $C_p$. The frames in the lower rows are data-transformed components of $C'_f$.}
	\label{trainSRT}
\end{figure}

\subsection{AE architecture}
The AE in our network aims to learn prototypical features of normal events and produce an output frame based on those features. Its main task is to predict $\mathbf{P_t}$—the frame coming after $\mathbf{C_f}$—from an input frame cuboid $\mathbf{C'_f}$.  Therefore, it is necessary to learn the temporal features as well as the spatial features to generate the frame with fine quality. The architecture of our model follows that of U-Net~\cite{ronneberger2015u}, in which the skip connections between the encoder and the decoder boost generation ability by preventing gradient vanishing and achieving information symmetry. The encoder consists of a stack of three-layer blocks that reduce the resolution of the feature map. We employ 3D convolution~\cite{tran2015learning} to embed the temporal factor learning in our model. Specifically, the first block consists of one convolutional layer and one activation layer. The second and the last blocks are identical in structure: convolutional, batch normalization, and activation layers. The kernel size is set to $3\times3\times3$ for all three layers. The decoder also consists of a stack of three-layer blocks and is symmetrical to the encoder except that the convolutional layers are replaced by deconvolutional layers to upscale the feature map. In addition, we use leakyReLU activation~\cite{maas2013rectifier} for the encoder and ReLU activation~\cite{nair2010rectified} for the decoder. \\
\indent Likewise, the architecture of our AE is very simple compared to other previous studies, especially methods that employ pre-trained feature extractors~\cite{luo2017revisit, sultani2018real}. In that the running time is generally dependent on the simplicity of the model architecture, our AE is well designed, considering the speed. 

\subsection{Objective function and normality score}
\noindent {\bf Prediction loss.} Our model is trained to minimize the prediction loss. We use the $L1$ distance (Eq. (\ref{L1})) and structural similarity index (SSIM)~\cite{wang2004image} loss (Eq. (\ref{ssim})) to measure the difference between the generated frame $\mathbf{P'_t}$ and the ground truth frame $\mathbf{P_t}$. The $L1$ distance and SSIM demonstrate the difference of frames at the pixel-level and similarity at the feature-level, respectively. The functions are as follows:
\vspace{-0.1cm}
\begin{equation}
	\label{L1}
	L_{p} (\mathbf{P'_t}, \mathbf{P_t})=| \mathbf{P'_t}, \mathbf{P_t}|
\end{equation}
\vspace{-0.5cm}
\begin{equation}
	\label{ssim}
	L_{f}(\mathbf{P'_t}, \mathbf{P_t}) = 1 - \frac{ (2 \mu _ { \mathbf{P'_t} } \mu _ { \mathbf{P_t} } + c _ { 1 } )(2 \sigma _ { \mathbf{P'_t}\mathbf{P_t} } + c _ { 2 } ) } { (2 \mu _ { \mathbf{P'_t} } ^ { 2 } \mu _ { \mathbf{P_t} } ^ { 2 } + c _ { 1 } )( \sigma _ { \mathbf{P'_t} } ^ { 2 } + \sigma _ { \mathbf{P_t} } ^ { 2 } + c _ { 2 } ) } ,
\end{equation}

\noindent where $\mu$ and $\sigma^ { 2 }$ denote the average and variance of each frame, respectively. Furthermore, $\sigma _ { \mathbf{P'_t}\mathbf{P_t} }$ represents the covariance. $c _ { 1 }$ and $c _ { 2 }$ denote variables to stabilize the division. Following the work of Zhao~\etal~\cite{zhao2016loss}, we exploit a weighted combination of the two loss functions in our objective function as shown in Eq. (\ref{eq1}). 
\begin{equation}
	\label{eq1}
	L_{pred} (\mathbf{P'_t}, \mathbf{P_t}) = \omega _ { p } L _ { p } (\mathbf{P'_t}, \mathbf{P_t}) + \omega _ { f } L _ { f } (\mathbf{P'_t}, \mathbf{P_t})
\end{equation}
$\omega _ { p }$ and $\omega _ { f }$ are the weights controlling the contribution of $L_{p}$ and $L_{f}$, respectively. Consequently, our model is urged to generate outputs that resemble the ground truth frames at both the pixel and feature levels.

\begin{table*}[!ht]
	\centering
	\resizebox{1.7\columnwidth}{!}{%
		\begin{tabularx}{\textwidth}{c|c|c|c|c|c|c}
			\hline \hline
			& Method & FPS & Prediction-based & \multirow{2}{*}{}{CUHK Avenue~\cite{lu2013abnormal}} & Shanghai Tech~\cite{luo2017revisit} & UCSD Ped2~\cite{mahadevan2010anomaly}\\
			\hline \hline
			\multirow{14}{*}{\rotatebox{90}{w/ pre-trained module\hspace{0.1cm}}}
			&StackRNN  \cite{luo2017revisit} & 50 & & 81.7 & 68.0 & 92.2 \\
			&FFP  \cite{Liu_2018_CVPR} & 133$^\dagger$ & {\ding{51}} & 85.1 & 72.8 & 95.4 \\
			
			&AD  \cite{ravanbakhsh2019training}  & 2 & & - & - & 95.5 \\
			&AMC  \cite{Nguyen_2019_ICCV} &  & {\ding{51}} & 86.9 & - & 96.2 \\
			&MemAE  \cite{gong2019memorizing} & 42$^\dagger$ & & 83.3 & 71.2 & 94.1 \\
			&DummyAno \cite{ionescu2019object} & 11 & & {87.4}$^\star$ & {78.7}$^\star$ & {94.3}$^\star$ \\
			&AnoPCN  \cite{ye2019anopcn} & 10&{\ding{51}} &  86.2  & 73.6 &  96.8  \\
			&GCLNC  \cite{zhong2019graph} & \textbf{150} & &  -  & \textbf{84.1} &  92.8  \\
			&GMVAE  \cite{fan2020video} & \underline{120} & &  83.4  & - &  92.2  \\
			&VECVAD  \cite{yu2020cloze} &5 & &  89.6  &  74.8  &  97.3  \\
			&FewShotGAN \cite{lu2020few} & &{\ding{51}} & 85.8 & 77.9 &96.2\\
			&AMmem  \cite{cai2021appearance} & & &  86.6  &  73.7  &  96.6  \\
			&MTL \cite{georgescu2021anomaly}& 21 & & \underline{91.5}$^\star$ & {82.4}$^\star$ & \underline{97.5}$^\star$ \\
			&BackAgnostic \cite{Georgescu_2021} & 18 & &  \textbf{92.3}  &  \underline{82.7}  &  \textbf{98.7} \\
			\hline
			
			\multirow{11}{*}{\rotatebox{90}{w/o pre-trained module \hspace{0.1cm}}}&150Matlab  \cite{Lu_2013_ICCV} & \underline{150}  & &  80.9  & - & -  \\
			&ConvAE  \cite{hasan2016learning} & & &  70.2  &  60.9  &  90.0  \\
			&HybridAE  \cite{nguyen2019hybrid} & & &  82.8  & - &  84.3  \\
			&CVRNN  \cite{lu2019future} & &{\ding{51}} &  85.8  & - &  96.1  \\
			&IntegradAE  \cite{tang2020integrating} &  30  &{\ding{51}} &  83.7  &  71.5  &  96.2  \\
			&MNAD  \cite{park2020learning} &  78$^\dagger$  &{\ding{51}}  &  \textbf{88.5}  &  70.5  &  \underline{97.0}  \\
			&MNAD  \cite{park2020learning} &  56$^\dagger$  &  &  82.8  &  69.8  &  90.2  \\
			&CDDAE \cite{chang2020clustering} &32 & & \underline{86.0} & \textbf{73.3} & 96.5 \\
			&OG  \cite{zaheer2020old}  & & & - & - &  \textbf{98.1}  \\
			\cline{2-7}
			&Baseline &  \textbf{195}  &{\ding{51}}  &  83.2  &  72.1  &  95.7  \\
			&Ours & \textbf{195}  &{\ding{51}}  &  85.3  &  \underline{72.2}  &  96.3  \\
			\hline \hline
			
		\end{tabularx}
	}
	\vspace{0.2cm}
	\caption{Frame-level AUC scores (\%) of the state-of-the-art methods versus our architecture trained with patch anomaly generation phase. For a fair comparison, like all other papers, the $\star$ marked scores are the micro-AUC performances taken from~\cite{georgescu2021Github,Georgescu_2021}. The FPS values are based on the figures mentioned in each paper, and the ones with $\dagger$ denote FPS computed in our re-implementation, conducted on the same device and environment as our model for a fair comparison. The top two results in each category are marked with \textbf{bold} and \underline{underline}.}
	\vspace{-0.2cm}
	\label{t2}
\end{table*}

\vspace{0.3em}
\noindent {\bf Frame-level anomaly detection.}
When detecting anomalies in the testing phase, we adopt the peak signal to noise ratio (PSNR) as a score to estimate the abnormality of the evaluation set. We obtain this value between the predicted frame at the $t ^ { th }$ period $\mathbf{P'_t}$ and the ground truth frame $\mathbf{P_t}$:
\begin{equation}
	PSNR(\mathbf{P'_t}, \mathbf{P_t}) = 10\log_{10} \frac{\max (\mathbf{P '_t})}{\|\mathbf{P'_t}-\mathbf{P_t}\|_2^2/N},
\end{equation}
where N denotes the number of pixels in the frame. Our model fails to generate when $\mathbf{P_t}$ contains abnormal events, resulting in a low value of PSNR and vice versa. Following the method of many related studies~\cite{chong2017abnormal, gong2019memorizing, hasan2016learning, lee2018stan, Liu_2018_CVPR,luo2017revisit, park2020learning, ravanbakhsh2019training}, we define the final normality score $S_t$ by normalizing $PSNR(\mathbf{P'_t}, \mathbf{P_t}$) of each video clip to the range $[0, 1]$. 

\begin{equation}
	\label{score}
	S_t = \frac{PSNR(\mathbf{P'_t}, \mathbf{P_t}) - \min PSNR(\mathbf{P'_t}, \mathbf{P_t})} {\max PSNR(\mathbf{P'_t}, \mathbf{P_t}) - \min PSNR(\mathbf{P'_t}, \mathbf{P_t})},
\end{equation}
Therefore, our model is capable of discriminating between normal and abnormal frames using the normality score of Eq. (\ref{score})

\section{Experiments}

\subsection{Implementation details}
We implement all of our experiments with PyTorch~\cite{paszke2017automatic}, using a single Nvidia GeForce RTX 3090. Our model is trained using Adam optimizer~\cite{kingma2014adam} with a learning rate of 0.0002. Additionally, a cosine annealing scheduler~\cite{loshchilov2016sgdr} is used to reduce the learning rate to 0.0001. We train our model for 20 epochs on the Avenue dataset~\cite{lu2013abnormal} and Ped2 dataset~\cite{mahadevan2010anomaly} and five epochs on the ShanghaiTech dataset~\cite{luo2017revisit}. The number of input frames $n$ is empirically set to 5. We load frames in gray scale in order to improve the speed and efficiency. Then we resize the frames to $240\times360$, and normalize the intensity of pixels to $[-1, 1]$. In addition, we add random Gaussian noise to the training input where the mean is set to 0 and the standard deviation is chosen randomly between 0 and 0.03. Furthermore, we set $W_p$ and $H_p$ to 60. The batch size is 4 during training. Optimal weights for the loss function in Eq. (\ref{eq1}) are empirically measured as $\omega _ { p } = 0.25$ and $\omega _ { f } = 0.75$.
\\

\noindent {\bf Evaluation metric.} We adopt the area under curve (AUC) of the receiver operating characteristic (ROC) curve obtained from the frame-level scores and the ground truth labels for the evaluation metric. This metric is used in most studies~\cite{chang2020clustering, georgescu2021anomaly, Georgescu_2021, lu2020few, luo2017revisit, nguyen2019hybrid, park2020learning, yu2020cloze, zaheer2020old} on video anomaly detection. Some works also report the localizing performance by adopting the pixel-level AUC. However, according to Ramachandra~\etal~\cite{9271895}, this criterion is a flawed metric because the results can be artificially improved by using some expedient tricks. Furthermore, this metric does not penalize the false positive detection within the true positive frames, meaning that the corresponding results are actually unsuitable for representing the spatial detecting performance. Therefore, new metrics called the Track-Based Detection Rate (TBDR) and the Region-Based Detection Rate (RBDR)~\cite{Ramachandra_2020_WACV} were proposed recently to replace the pixel-level AUC. However, the official implementation data have yet to be released. Hence, we only consider the temporal evaluation in this paper. \\
\indent The baseline model, mentioned throughout the following sections, denotes our model without the patch anomaly generation phase. Since the first five frames of each clip cannot be predicted, they are ignored in the evaluation, following~\cite{Liu_2018_CVPR, park2020learning, tang2020integrating}.

\subsection{Datasets}
We evaluate our model with three datasets which are all acquired from the real-world scenarios. 
\vspace{0.2em}

{\noindent \bf CUHK Avenue~\cite{lu2013abnormal}.}
This dataset captures an avenue at a campus. It consists of 16 training and 21 testing clips. Training clips contain only normal events and testing clips contain a total of 47 abnormal events such as running, loitering, and throwing objects. The frame resolution is $360 \times 640$, all in RGB scale. The size of people is inconsistent due to the camera angle. Furthermore, the camera is kept fixed most of the time. However, a subtle shaking is recorded briefly in the test set. 

\vspace{0.2em}
{\noindent \bf UCSD Ped2~\cite{mahadevan2010anomaly}.}
The UCSD Ped2 dataset~\cite{mahadevan2010anomaly} is acquired from a pedestrian walkway by a fixed camera from a long distance. The training and the testing sets consist of 16 and 12 clips, respectively. Anomalies in the testing clips are non-pedestrian objects, for instance, bikes, cars, and skateboards. The frames are in gray scale with a resolution of $240 \times 360$. 

\vspace{0.2em}
{\noindent \bf ShanghaiTech Campus~\cite{luo2017revisit}.}
Unlike the others, this dataset contains multi-scene anomalies and is the most complex and largest dataset. It is acquired from 13 different scenes. There are 330 training videos and 107 testing videos where non-pedestrian objects ({\it e.g.,} cars, bikes) and aggressive motions ({\it e.g.,} brawling, chasing) are annotated as anomalies. Each frame is captured with $480 \times 856$ RGB pixels. 

\subsection{Experimental results}
\label{section4}
\noindent {\bf Impact of patch anomaly generation phase.}
Table~\ref{augmentationAblation} shows the impact of our patch anomaly generation estimated on Avenue~\cite{lu2013abnormal} and Ped2~\cite{mahadevan2010anomaly}. The results include five different conditions: (1) using only TMT, (2) using only SRT, (3) randomly applying TMT or SRT but with all patches rotated as a chunk in the same direction for SRT, where $\delta_t=\delta_{t+1}= \dots =\delta_{n-1}$ (represented as SRT* in Table~\ref{augmentationAblation}), (4) randomly applying TMT or SRT with varying directions for each patch, and (5) applying both TMT and SRT to the selected $\mathbf{C_p}$. From the results, it appears that SRT has a greater contribution than TMT to the detection performance. This is because our SRT rotates each patch randomly in varying directions resulting in generating anomalies in the motion as well as the appearance. 

\begin{table}
	\centering
	\begin{adjustbox}{width=0.8\linewidth}
		\begin{tabular}{ c|c|c|c } 
			\hline
			Method & Avenue~\cite{lu2013abnormal} & ST~\cite{luo2017revisit} & Ped2~\cite{mahadevan2010anomaly} \\
			\hline\hline
			Baseline & 83.2 & 72.1 & 95.7 \\
			TMT &  83.0  &  72.2  &  95.1  \\ 
			SRT &  85.0  & \textbf{72.4} & 96.0 \\ 
			TMT $\bigvee$ SRT* &  84.5  &  72.1  &  95.2  \\ 
			TMT $\bigvee$ SRT& \textbf{85.3} &  72.2  & \textbf{96.3} \\
			TMT $\bigwedge$ SRT&  84.6  &  72.2  &  96.2  \\
			\hline
		\end{tabular}
	\end{adjustbox} 
	\vspace{0.3cm}
	\caption{We demonstrate the impact of our patch anomaly generation by ablation studies on CUHK Avenue~\cite{lu2013abnormal}, ShanghaiTech (ST)~\cite{luo2017revisit}, and Ped2~\cite{mahadevan2010anomaly}. We present frame-level AUC (\%) of experiments on 5 variations: using only TMT, using only SRT, randomly selecting between TMT and single directional SRT (indicated as SRT*), randomly selecting between TMT and SRT, and using both the TMT and SRT.}
	\label{augmentationAblation}
\end{table}
\vspace{0.3em}

\begin{figure}[!t]
	\begin{center}
		\vspace{-0.5cm}
		\includegraphics[width=0.7\linewidth]{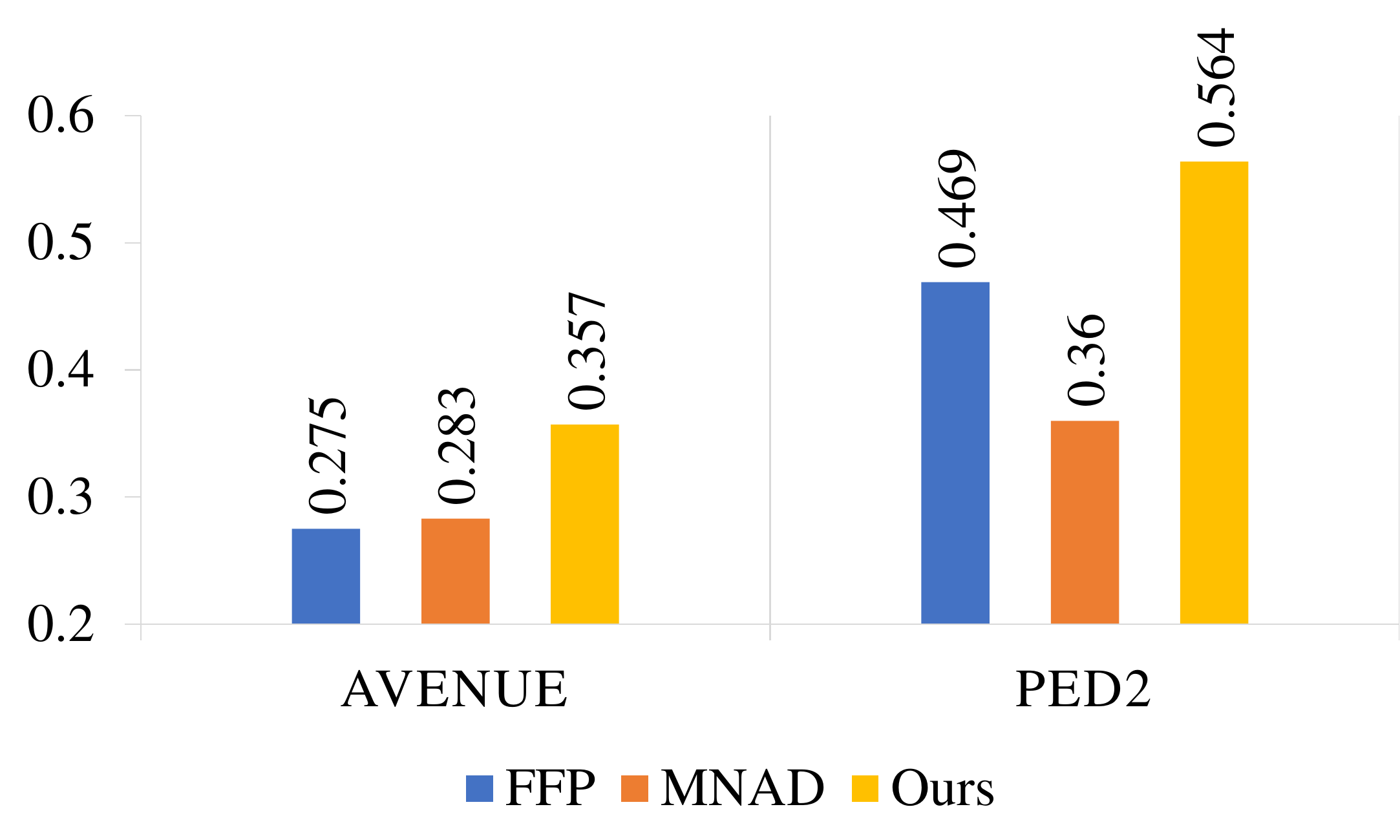}
	\end{center}
	\vspace{-0.5cm}
	\caption{Following the work of Liu~\etal~\cite{Liu_2018_CVPR}, we compare our work with FFP~\cite{Liu_2018_CVPR} and MNAD~\cite{park2020learning} by calculating the score gap between normal frames and abnormal frames on CUHK Avenue~\cite{lu2013abnormal} and UCSD Ped2~\cite{mahadevan2010anomaly}. The gap is obtained by averaging the scores of normal frames and those of abnormal frames and subtracting the two values. A higher gap represents a higher capacity for discriminating normal and abnormal frames.}
	\label{scoregap}
\end{figure}

\noindent {\bf Performance comparison with existing works.}
We compare the frame-level AUC of our model with those of non-prediction-based methods~\cite{hasan2016learning,luo2017revisit,ravanbakhsh2017abnormal, sultani2018real, ravanbakhsh2019training, nguyen2019hybrid, gong2019memorizing, ionescu2019object, park2020learning, yu2020cloze, zaheer2020old, georgescu2021anomaly, Georgescu_2021} and prediction-based methods~ \cite{Liu_2018_CVPR, tang2020integrating, Nguyen_2019_ICCV, park2020learning}. From Table~\ref{t2}, we find that our method achieves competitive performance on the three datasets with a very high temporal rate. Among the prediction-based methods, we exceed IntergadAE~\cite{tang2020integrating} in all datasets and show superior results especially in the Ped2 dataset~\cite{mahadevan2010anomaly}. Note that our model performs at par with other models without any additional modules whereas several other prediction-based models~\cite{Liu_2018_CVPR, tang2020integrating, Nguyen_2019_ICCV} employed pre-trained optical flow networks to estimate the motion features. Among the non-prediction-based networks, Georgescu~\etal~\cite{georgescu2021anomaly} achieved superior performance by combining self-supervised learning with a pre-trained object detector.\\
\indent Furthermore, we conduct a score gap comparison, inspired by Liu~\etal~\cite{Liu_2018_CVPR} to present the discriminating capacity of our model. Fig.~\ref{scoregap} shows that our model achieves higher gaps than FFP~\cite{Liu_2018_CVPR}—a prediction network boosted with optical flow loss and generative learning, and MNAD~\cite{park2020learning}—a prediction method that reads and updates memory items from a memory module. This demonstrates the effectiveness of our patch anomaly generation phase by the fact that the score distributions of normal and abnormal frames are significantly far apart from each other. 

\begin{figure}[!t]
	\begin{center}
		\includegraphics[width=0.8\linewidth]{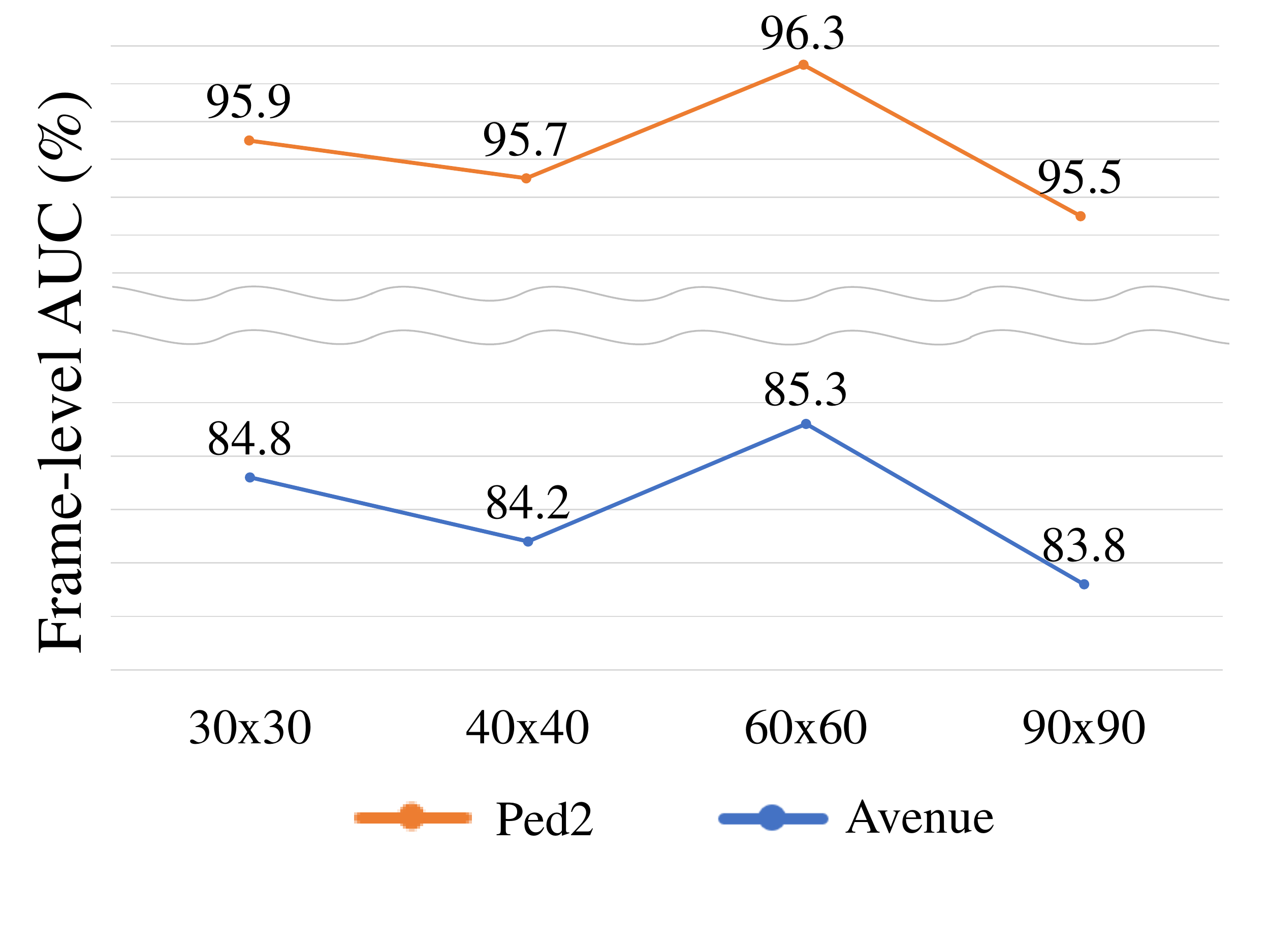}
	\end{center}
	\vspace{-0.5cm}
	\caption{Results of ablation studies on patch size.}
	\label{patchsizeAblation}
\end{figure}
\vspace{0.3em}

\vspace{-0.3em}
\noindent {\bf Running time.}
Our model boasts an astonishing speed of 195 frames per second (FPS). This rate is computed using UCSD Ped2~\cite{mahadevan2010anomaly} test set with a single Nvidia GeForce RTX 3090 GPU. We obtain this by averaging the entire time consumed in both frame generation and anomaly prediction. To our knowledge, it is far faster than any other previous works. We show a fair comparison with other networks in Table~\ref{t2}. We re-implemented networks that distributed official codes in public on the same device and environment used for our network. The FPS for these is marked with $\dagger$ in the table. We copied the figures mentioned in each paper for methods without publicly distributed codes. Note that our work is nearly 30 \% faster than the second-fastest ones~\cite{zhong2019graph, Lu_2013_ICCV}. Moreover, we computed the number of trainable parameters as proof of 195 FPS. Its value for our model is 2.15 M whereas it is 15.65 M for MNAD\cite{park2020learning} with 67 FPS and 14.53 M for FFP\cite{Liu_2018_CVPR} with 25 FPS. Our network is remarkably cheaper in computation than the compared methods.

\begin{figure}[t]
	\centering
	\subfloat[CUHK Avenue]{\includegraphics[width=0.95\linewidth]{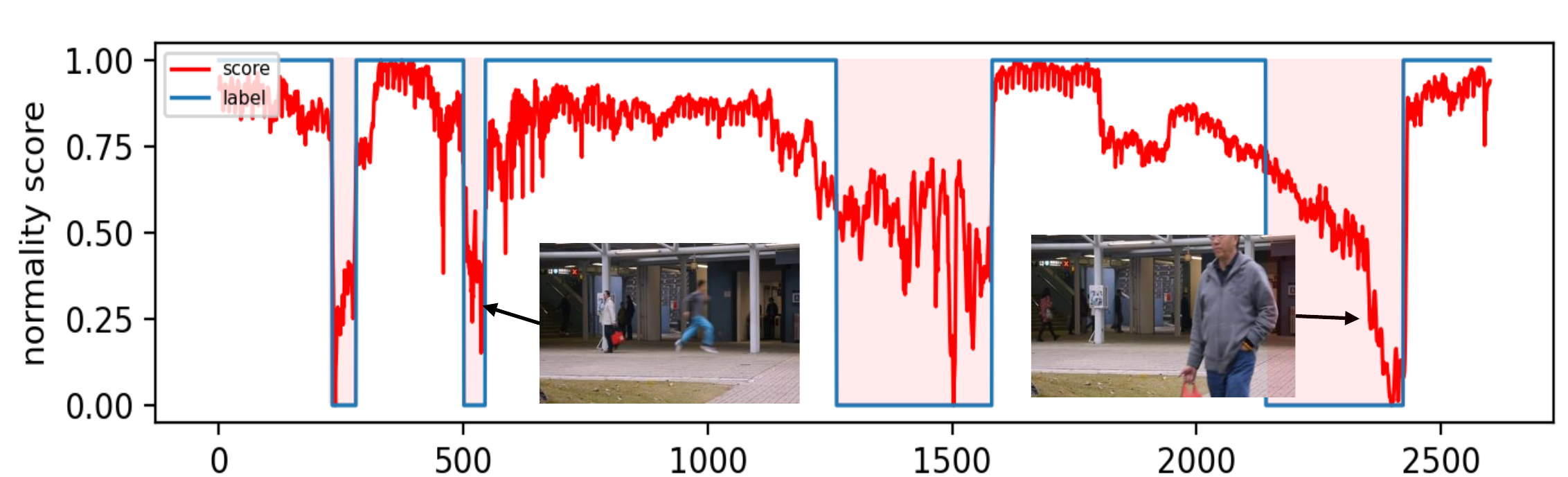}
	}
	\\
	\vspace{-0.8em}
	\subfloat[ShanghaiTech]{\includegraphics[width=0.95\linewidth]{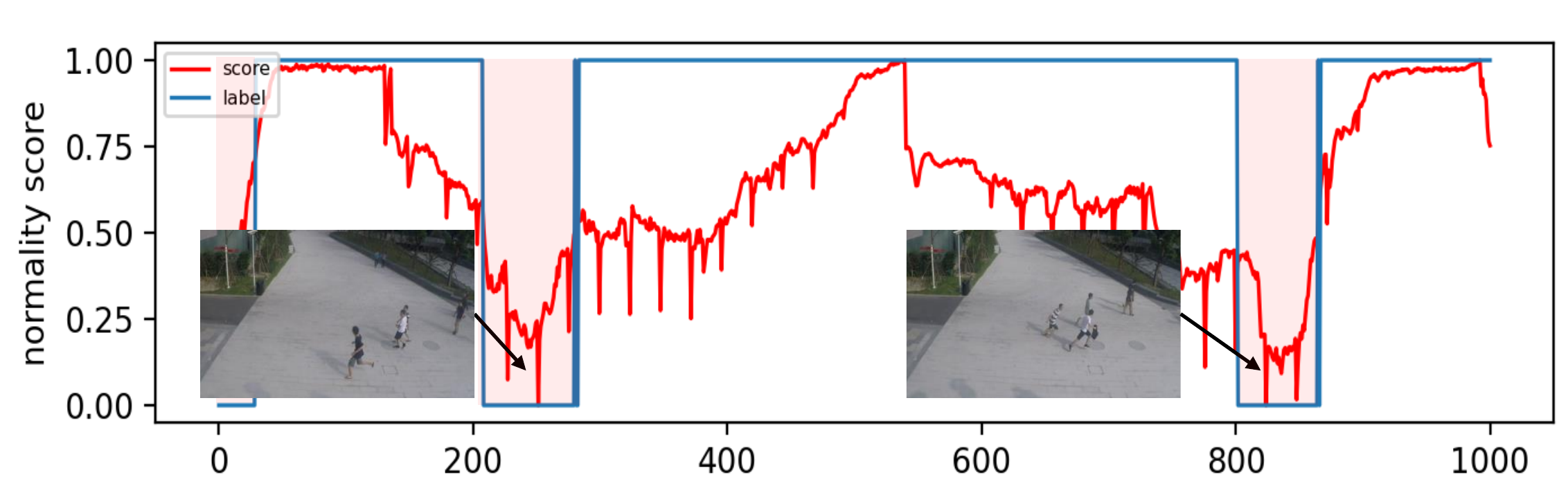}
	}\\
	\vspace{-0.8em}
	\subfloat[UCSD Ped2]{\includegraphics[width=0.95\linewidth]{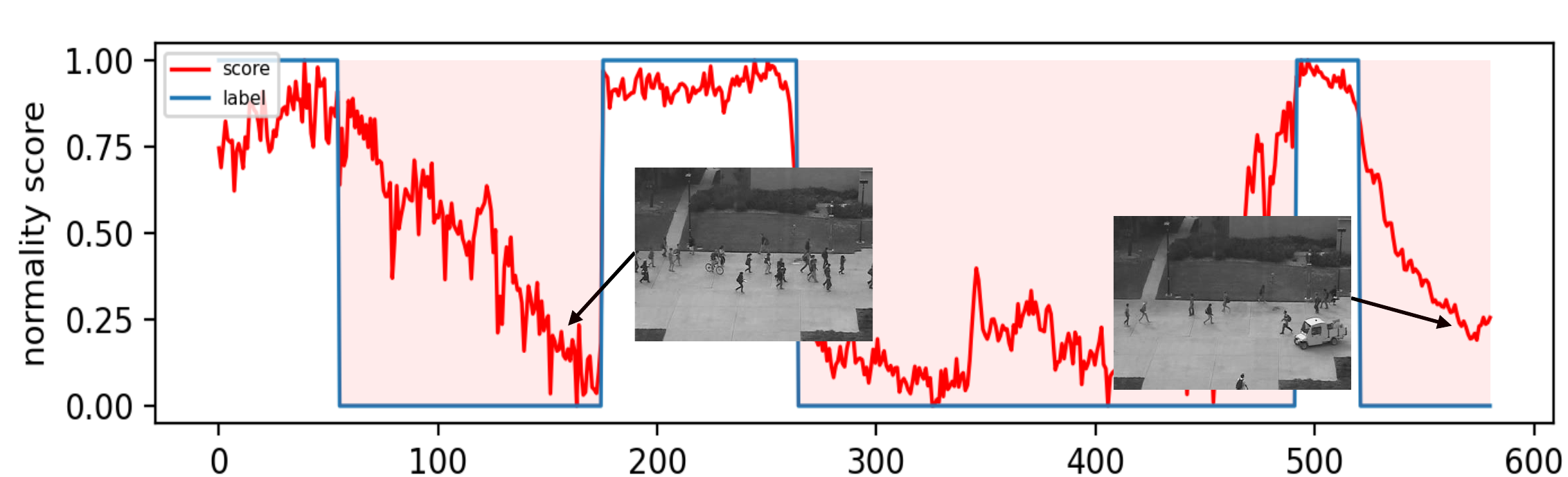}
	}\\
	\vspace{0.5em}
	\caption{Score plot from evaluation. The red and blue lines denote $S_t$ and labels, respectively. Labels are 0 when frames are abnormal. (A) is obtained from Avenue~\cite{lu2013abnormal}. Running, throwing a bag, and moving in the wrong direction are well detected. (B) is obtained from ShanghaiTech~\cite{luo2017revisit}. Chasing and running are detected as anomalies. (C) is obtained from Ped2~\cite{mahadevan2010anomaly} where the captured anomalies are bicycles and a car.}
	\label{scoreplot}
\end{figure}

\vspace{0.3em}
\noindent {\bf Ablation studies on patch size.} 
Fig.~\ref{patchsizeAblation} shows the result of ablation experiments that we conducted on the Avenue~\cite{lu2013abnormal} and Ped2~\cite{mahadevan2010anomaly} to observe the effect of the patch size. The patch size determines the smallest unit to be focused on by the AE. In all experiments of these ablation studies, only the size of the patch is changed between $30 \times 30$, $40 \times 40$, $60 \times 60$, and $90 \times 90$, while the frame resolution remains fixed at $240 \times 360$. It means that a comparably small region is captured in a $\mathbf{C_f}$ with the size of $30 \times 30$, and a large region is captured in a $\mathbf{C_f}$ with the size of $90 \times 90$. Our network shows the lowest accuracy when the patch size is $90 \times 90$, which is more than 10 percent of the frame size. When the patch is considerably large, the model focuses on larger movements than smaller ones. Abnormal conditions usually occur in small parts, hence, lower performance is observed in this case. 

\begin{figure}[!ht]
	\centering
	\subfloat[UCSD Ped2]{\includegraphics[width=0.95\linewidth]{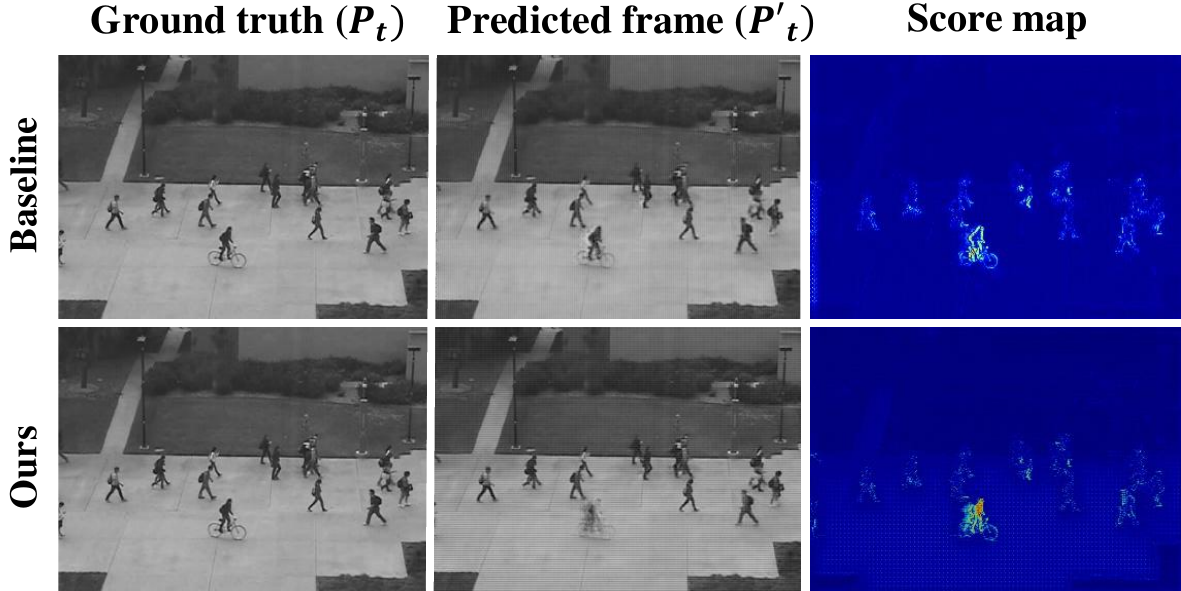}
	}
	
	\vspace{-0.3cm}
	\subfloat[CUHK Avenue]{\includegraphics[width=0.95\linewidth]{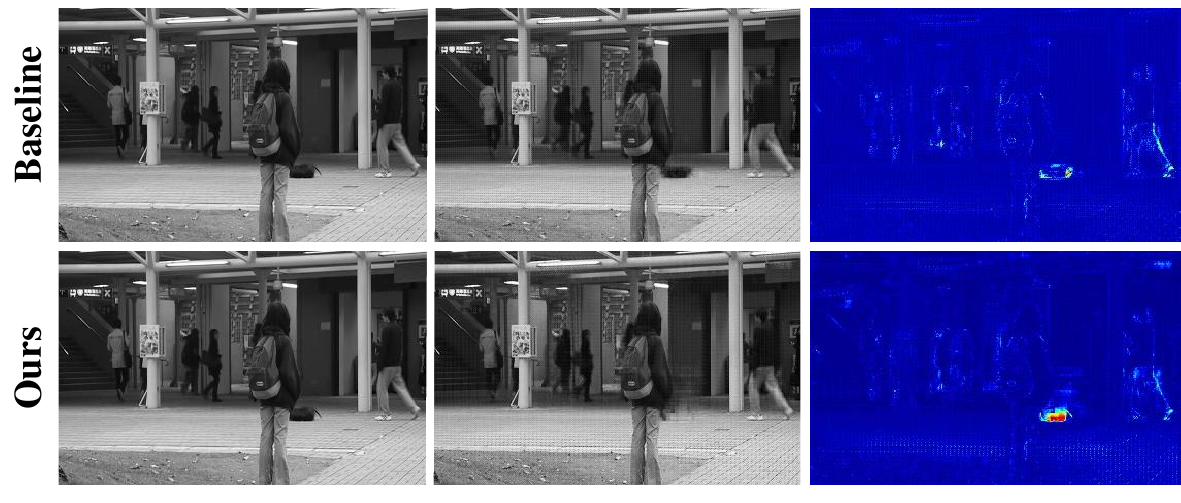}
	}
	\vspace{-0.3cm}
	\subfloat[ShanghaiTech]{\includegraphics[width=0.95\linewidth]{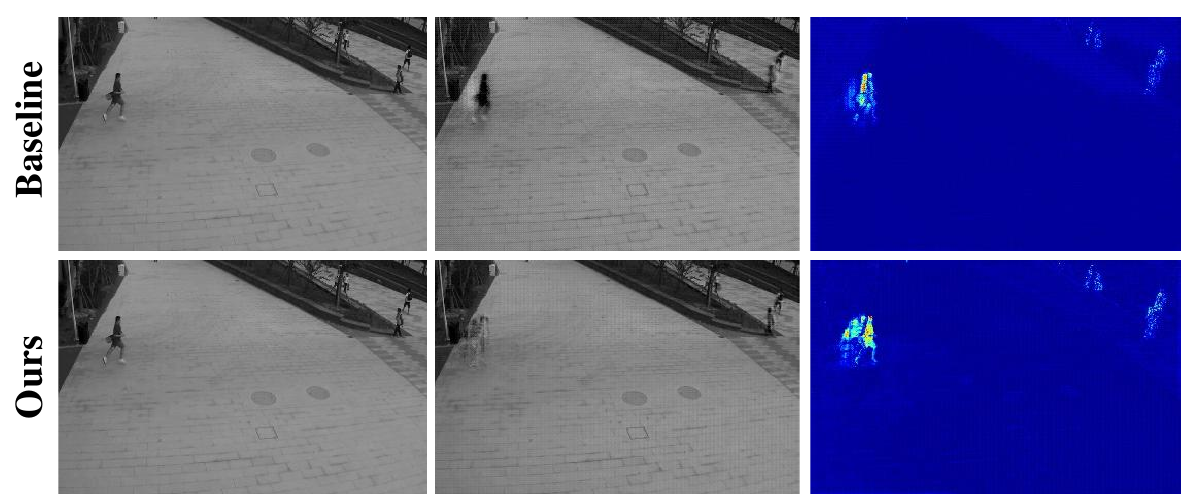}
	}
	\vspace{0.5em}
	\caption{Examples of predicted frames and difference maps compared to our baseline. Best viewed in color.}
	\label{output}
\end{figure}

\vspace{0.3em}
\noindent {\bf Qualitative results.}
We demonstrate the frame-level detecting performance of our model in Fig.~\ref{scoreplot}. From the figure, it can be observed that $S_t$ rapidly decreases when anomalies appear in the frames. Once the abnormal objects disappear, $S_t$ increases immediately.\\
\indent Furthermore, the pixel-level detecting capacity is observed in Fig.~\ref{output}. We present examples of predicted frames and the corresponding difference maps. Additionally, we emphasize the results by comparing each sample with those of our baseline model. In the example of Ped2~\cite{mahadevan2010anomaly}, the bicycle is the annotated anomaly, which is an unseen appearance. In Avenue~\cite{lu2013abnormal} and ShanghaiTech~\cite{luo2017revisit}, the annotated anomalies relate to motion: a man throwing a bag and a running person. The outputs generated by our model trained with the patch anomaly generation phase are significantly much blurrier than those of the baseline, validating the effectiveness of our transformation phase. Note that our model nearly erased the bag and the person in the examples of Avenue~\cite{lu2013abnormal} and ShanghaiTech~\cite{luo2017revisit}. This proves that our model does not simply infer abnormal objects by copying from the inputs, which is what the baseline model does. Moreover, for the ShanghaiTech dataset~\cite{luo2017revisit}, the difference map of our model shows a distinction in a larger region compared to that of the baseline. We observe that our model did not accept the motion in the input; it attempted to predict the trajectory of the runner as it as per the training. However, the baseline model generated a moderate copy of the input based on the given trajectory. 
\vspace{-0.2cm}
\section{Conclusion and Future Work}
\vspace{-0.2cm}
In this paper, we proposed a prediction network for video anomaly detection combined with a patch anomaly generation phase. We designed a light-weight AE model to learn the common spatio-temporal features of normal frames. The proposed method generated transformed frame cuboids as inputs, by applying SRT or TMT to a random patch cuboid within the frame cuboid. Our model was encouraged to pay attention to the appearance and motion patterns of normal scenes. In addition, we discussed the impact of the patch anomaly generation by conducting ablation studies. Furthermore, the proposed method achieved competitive performance on three benchmark datasets and performed at a very high speed, which is as important as the detection capacity in anomaly detection.\\
\indent 
Through the experimental results, we also have shown that our network is able to localize the anomalies. Since detecting temporal-wise anomalies is the most essential part and there is an inconsistency issue in pixel-level AUC evaluation~\cite{9271895}, we only considered the frame-level detection. With the newly proposed TBDR and RBDR metrics~\cite{Ramachandra_2020_WACV}, our future work will be verifying the fast localizing capability of our network.\\


\newpage

{\small
	\bibliographystyle{ieee_fullname}
	\bibliography{egbib}
}

\end{document}